\documentclass{article}
\usepackage{flowmem_arxiv,times}

\usepackage{amsmath,amsfonts,bm}

\def\eqref#1{equation~\ref{#1}}

\def\1{\bm{1}}

\DeclareMathAlphabet{\mathsfit}{\encodingdefault}{\sfdefault}{m}{sl}
\SetMathAlphabet{\mathsfit}{bold}{\encodingdefault}{\sfdefault}{bx}{n}

\usepackage[utf8]{inputenc}
\usepackage[T1]{fontenc}
\usepackage{hyperref}
\usepackage{url}
\usepackage{booktabs}
\usepackage{amsfonts}
\usepackage{amsmath}
\usepackage{amsthm}
\usepackage{bm}
\usepackage{graphicx}
\usepackage{microtype}
\usepackage[table]{xcolor}
\usepackage{xspace}
\usepackage{enumitem}
\usepackage[small]{caption}
\usepackage{subcaption}
\usepackage{multirow}
\usepackage{makecell}
\usepackage{flafter}
\usepackage{placeins}
\usepackage{wrapfig}
\usepackage{float}
\usepackage{algorithm}
\usepackage{algorithmic}

\newcommand{\method}{\textsc{FlowMem}\xspace}
\newcommand{\lastzero}{LaST$_0$\xspace}
\setlist[itemize]{leftmargin=1.3em,labelsep=0.45em,topsep=0.2em,itemsep=0.15em,parsep=0pt,partopsep=0pt}
\title{Recompose and Refine Latent Reasoning Flows\\
for Vision-Language-Action Models}

\renewcommand{\thefootnote}{\fnsymbol{footnote}}
\author{
\textbf{Hongyu Shi\textsuperscript{1} \quad
Sen Zhao\textsuperscript{2}\thanks{Corresponding authors: \texttt{zhaosen@cqupt.edu.cn}, \texttt{zhangx@cqupt.edu.cn}.} \quad
Zuyu Zhang\textsuperscript{2} \quad
Lifeng Shen\textsuperscript{3}} \\
\textbf{Ding Zou\textsuperscript{4} \quad
Xinyu He\textsuperscript{5} \quad
Xu Zhang\textsuperscript{2}\footnotemark[1] \quad
Qinghua Zhang\textsuperscript{1}} \\[0.35em]
{\normalfont\small \textsuperscript{1}School of Computer Science and Technology,} \\
{\normalfont\small Chongqing University of Posts and Telecommunications, Chongqing, China} \\
{\normalfont\small \textsuperscript{2}Academy of Advanced Interdisciplinary Studies,} \\
{\normalfont\small Chongqing University of Posts and Telecommunications, Chongqing, China} \\
{\normalfont\small \textsuperscript{3}School of Artificial Intelligence,} \\
{\normalfont\small Chongqing University of Posts and Telecommunications, Chongqing, China} \\
{\normalfont\small \textsuperscript{4}ZTE Corporation, China} \\
{\normalfont\small \textsuperscript{5}Towngas, China}
}

\hypersetup{
  hidelinks,
  pdftitle={Recompose and Refine Latent Reasoning Flows for Vision-Language-Action Models},
  pdfauthor={Hongyu Shi, Sen Zhao, Zuyu Zhang, Lifeng Shen, Ding Zou, Xinyu He, Xu Zhang, Qinghua Zhang},
  pdfsubject={FlowMem preprint},
  pdfkeywords={Vision-Language-Action Models, Latent Reasoning, Memory, FlowMem}
}

\begin{document}
\maketitle
\renewcommand{\thefootnote}{\arabic{footnote}}
\setcounter{footnote}{0}

\begin{abstract}
Latent reasoning enables vision-language-action (VLA) models to transform
multimodal observations into task-relevant internal states before generating
continuous robot actions. While existing methods learn to generate or refine
such states for each policy query, they discard successful reasoning after
execution and therefore reconstruct similar computation from scratch. We
present \emph{Reasoning and Flow Memory} (\method), a unified VLA model that
turns successful latent computation into reusable reasoning experience.
Rather than appending a fixed retrieved context, \method dynamically retrieves
and recomposes compatible latent fragments as the embodied context evolves,
forming a reasoning route that follows the temporal structure and progress of
successful computation. The route is then refined using current visual and
proprioceptive evidence before it conditions action generation. Experiments on
RoboMME and LIBERO-Plus show that \method attains 48.0\% and 77.3\%
success---1.7 and 4.1 points above memory-free policies,
respectively---demonstrating the value of reusing successful latent computation
for closed-loop VLA control.
\end{abstract}

\section{Introduction}
\label{sec:introduction}

Vision-language-action (VLA) models provide a unified interface between multimodal perception, language understanding, and continuous robot control. Recent work has moved beyond direct observation-to-action prediction by introducing intermediate reasoning before action generation. Explicit approaches verbalize embodied plans or predict future visual goals, while latent-reasoning VLAs internalize such computation into compact continuous states \citep{zawalski2024ecot,zhao2025cotvla,huang2025thinkact}. In particular, LaRA-VLA combines textual and visual prediction in latent reasoning slots, LaST$_0$ learns a spatio-temporal latent chain-of-thought from future physical states, and RD-VLA improves action prediction through recurrent latent computation \citep{bai2026laravla,liu2026last0,tur2026rdvla}. These advances establish latent reasoning as an effective intermediate representation between perception and action, avoiding the latency and representational bottlenecks of lengthy explicit reasoning traces.

\begin{wrapfigure}{R}{0.52\textwidth}
    \centering
    \vspace{-4pt}
    \includegraphics[width=0.50\textwidth]{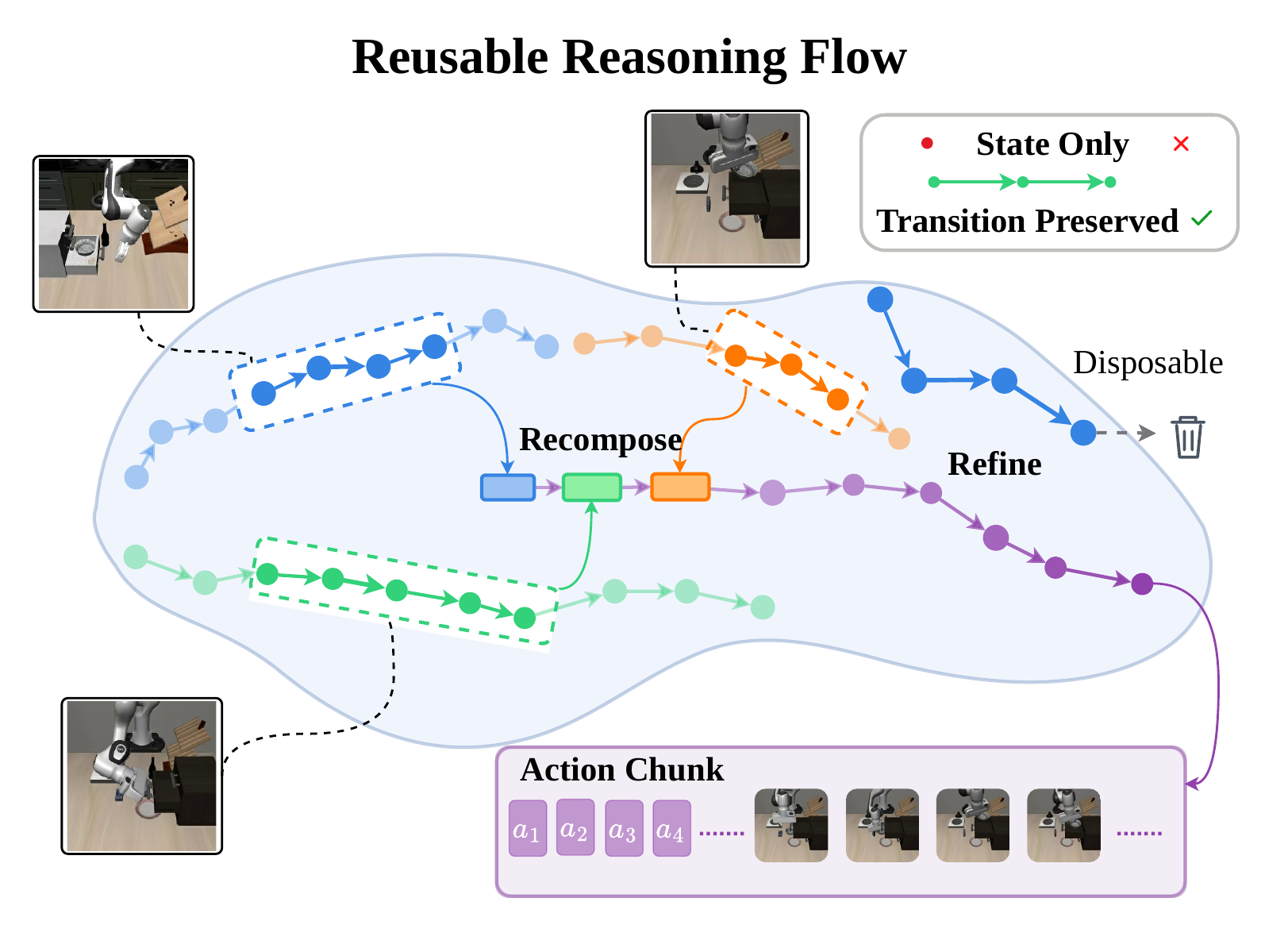}
    \caption{Comparison between transient latent reasoning and the proposed reusable reasoning-flow formulation.}
    \label{fig:motivation}
    \vspace{-8pt}
\end{wrapfigure}

However, latent reasoning remains largely transient. As illustrated in Fig.~\ref{fig:motivation}, a VLA constructs an intermediate trajectory from the current observation, uses it to generate actions, and discards the computation afterward. Even when the robot later encounters a familiar task stage, the successful reasoning that previously connected perception to control cannot be directly reused. Memory-augmented VLAs address a related but different limitation by preserving textual notes, perceptual-cognitive evidence, retrieved observations, or recurrent policy states \citep{haresh2026notestoself,shi2026memoryvla,cherepanov2026muvla,li2026trmvla}. Such memories enrich the historical context available to the policy, but they do not retain how latent reasoning itself evolved toward successful control. This leaves a central gap between \emph{reasoning before acting} and \emph{learning from how successful reasoning unfolded}.

Reusing latent reasoning is fundamentally different from retrieving a similar embedding. Its utility depends on task progress: a nearby state carries neither the direction of computation nor its unfinished portion. We therefore represent successful embodied reasoning as a \emph{latent reasoning flow}, an ordered, progress-aware trajectory of latent computation.

This perspective raises a new question: \emph{can a VLA construct its current reasoning process from previously successful latent computation, rather than regenerating it from scratch?} Figure~\ref{fig:motivation} exposes the three requirements behind this question. First, retrieval must identify a compatible \emph{transition fragment}, since a nearby latent point alone carries neither progress nor direction. Second, no single episode need cover the unfinished task, requiring complementary fragments to be recomposed without destroying their internal order. Third, the composed route is only prior computation: differences in object pose, layout, and robot state must be corrected by the current observation before the route can condition control.

Building on this formulation, we propose \emph{Reasoning and Flow Memory} (\method), a unified VLA model that transforms successful latent computation into reusable reasoning experience. \method comprises four coordinated components. The \emph{Embodied Context Encoder} represents the current observation, robot state, and instruction. The \emph{Reasoning Flow Memory} organizes successful latent states into ordered and grounded experience rather than isolated memory vectors. Conditioned on the current embodied context, the \emph{Recompose-and-Refine Reasoning Expert} retrieves anchor-aligned suffixes, constructs a compatible multi-source route, and refines it for the current scene. Finally, the \emph{Memory-Adaptive Action Expert} injects the refined route through the VLA's native latent interface to produce executable action chunks. All four components communicate through a shared latent representation and form a single perception--reasoning--action model.

Recomposition determines which prior transitions support the unfinished task, while refinement determines how the resulting route should change under current embodied evidence. The route is compressed into fixed reasoning tokens and corrected by a fixed-depth residual block before action generation.

We evaluate \method on RoboMME \citep{dai2026robomme} and LIBERO-Plus \citep{fei2026liberoplus}. These benchmarks jointly assess memory-dependent manipulation across temporal, spatial, object, and procedural requirements, as well as the reuse of latent reasoning under systematic distribution shifts. Extensive experiments compare \method with representative latent-reasoning and memory-augmented VLAs, while controlled memory interventions examine how its behavior changes with ordered flow recomposition and current-context refinement. Beyond task success, we further evaluate reasoning reuse and computational efficiency.

Our contributions are summarized as follows:
\begin{itemize}
    \item We introduce latent reasoning flows as a reusable representation of successful embodied computation, capturing how internal reasoning evolves across task progress rather than reducing experience to isolated vectors or historical policy context.

    \item We propose \method, a unified VLA model whose Recompose-and-Refine Reasoning Expert constructs task-specific reasoning from multiple experiences and refines it through current-context residual refinement before action generation.

    \item Empirical results on RoboMME and LIBERO-Plus indicate that treating
    successful latent computation as ordered reasoning experience can improve
    closed-loop VLA control, with gains of 1.7 and 4.1 points,
    respectively.
\end{itemize}

\section{Preliminaries}
\label{sec:preliminaries}

\subsection{Vision-Language-Action Models with Latent Reasoning}
Given an RGB observation $o_t$, robot state $s_t$, and language instruction
$l$, a vision-language-action (VLA) model predicts a chunk of $H$ future
actions $\mathbf a_{t:t+H-1}\in\mathbb R^{H\times d_a}$. Each action commonly
specifies end-effector translation, rotation, and gripper control. Modern VLAs
often generate continuous actions with diffusion or flow-matching objectives,
while latent-reasoning VLAs additionally introduce an intermediate state
$\mathbf Z_t\in\mathbb R^{L\times d}$ between multimodal context and action
generation \citep{liu2026last0,bai2026laravla,tur2026rdvla}:
\begin{equation}
    \mathbf Z_t = R_\theta(o_t,s_t,l),
    \qquad
    \hat{\mathbf a}_{t:t+H-1}=A_\theta(o_t,s_t,l,\mathbf Z_t).
\end{equation}
Here, $R_\theta$ denotes the native latent-reasoning module, $A_\theta$ the
action expert, and $\mathbf Z_t$ the model's internal task-relevant computation,
rather than an explicit textual plan, subgoal label, or action sequence.
Existing methods generate or iteratively refine this state for the current
policy query; the resulting computation is normally discarded after action
prediction.

\subsection{Problem Formulation}
\label{sec:problem_formulation}
We consider language-conditioned robotic manipulation as a partially observed
sequential decision problem. An episode is
$\zeta=(l,\{o_t,s_t,\mathbf a_t\}_{t=1}^{T},y)$, where $y\in\{0,1\}$ denotes
task success. Let $\mathcal D^+=\{\zeta_i\mid y_i=1\}$ contain successful
training episodes and let $\mathcal M=\Phi(\mathcal D^+)$ be reusable memory
constructed from their latent reasoning traces. Our goal is to learn a unified
VLA policy
\begin{equation}
    (\mathbf Z_t,\hat{\mathbf a}_{t:t+H-1})
    =\Pi_\theta(o_t,s_t,l,\mathcal M)
\end{equation}
where $\Pi_\theta$ denotes the memory-conditioned VLA policy that uses compatible prior computation to improve expected task success under
the deployment distribution. Memory informs the current latent reasoning state,
while every predicted action remains conditioned on the current observation and
robot state.

At evaluation time, the policy may access only the current embodied context
and memory committed before the evaluated episode. Memory-source and evaluation episodes are disjoint; future frames,
rewards, success labels, simulator-privileged states, and retrieval from the
currently executing episode are unavailable. Under these constraints, the
problem is to reuse successful latent computation without
sacrificing closed-loop adaptation to the present physical state.

\section{The Proposed Model}
\label{sec:method}

\subsection{Overview}
Given an RGB observation $o_t$, robot state $s_t$, and language instruction $l$, a VLA policy predicts an action chunk $\hat{\mathbf a}_{t:t+H-1}$. We augment this mapping with a reasoning flow memory $\mathcal M$ that stores latent reasoning produced during successful interactions:
\begin{equation}
    \hat{\mathbf a}_{t:t+H-1}
    = \Pi_\theta(o_t,s_t,l,\mathcal M).
\end{equation}
where $\Pi_\theta$ denotes the memory-conditioned policy and $\mathcal M$ the
reasoning flow memory.
The central premise of \method is that latent reasoning is reusable computation. Instead of retrieving past observations or replaying past actions, the model retrieves ordered latent fragments that describe how successful reasoning evolved, recomposes compatible fragments into a route for the current task, and refines the query-local route tokens using the current embodied context.

As illustrated in Fig.~\ref{fig:architecture}, \method contains four components. The \emph{Embodied Context Encoder} represents the current observation, state, and instruction. The \emph{Reasoning Flow Memory} stores ordered latent fragments from successful source episodes together with compact retrieval and transition metadata. The \emph{Recompose-and-Refine Reasoning Expert} performs anchor-aligned fragment retrieval, route tokenization, and current-context refinement. Finally, the \emph{Memory-Adaptive Action Expert} injects the refined reasoning tokens through the native latent interface of the VLA and generates continuous actions.

\begin{center}
    \centering
    \makebox[\linewidth][c]{\includegraphics[width=\linewidth]{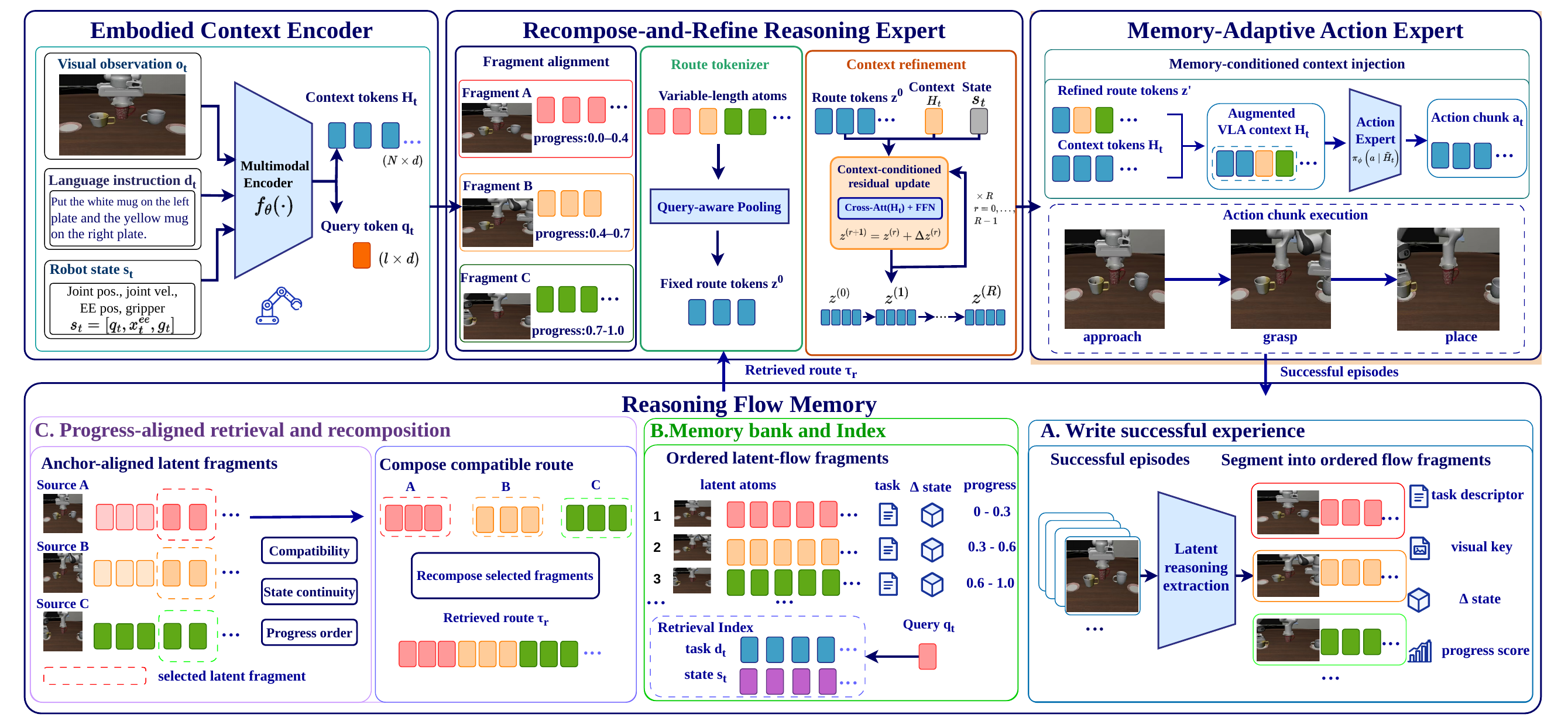}}
    \captionof{figure}{Overview of \method. Successful latent reasoning is stored as ordered flow fragments, retrieved and recomposed according to the current embodied context, and refined before conditioning action generation.}
    \label{fig:architecture}
\end{center}

\subsection{Embodied Context Encoder}
\label{sec:context_encoder}
The Embodied Context Encoder maps the current multimodal input to contextual tokens
\begin{equation}
    \mathbf H_t
    = E_\theta(o_t,s_t,l)
    \in\mathbb R^{N_c\times d}.
\end{equation}
where $N_c$ is the number of contextual tokens and $d$ is their latent width.
These tokens are used both by the action policy and by the memory pathway. For retrieval, we derive a compact visual semantic descriptor $\mathbf d_t$ from the observable context and pair it with a normalized robot-state vector $\bar{\mathbf s}_t$. The resulting query is
\begin{equation}
    \mathbf q_t=(\mathbf d_t,\bar{\mathbf s}_t,l).
\end{equation}
Only information available at the current policy query is used online. Future observations, rewards, and success labels are used solely to construct or supervise memory offline.

\subsection{Reasoning Flow Memory}
\label{sec:flow_memory}
For each successful source episode $i$, we record the latent reasoning states produced at successive policy queries,
\begin{equation}
    \mathcal Z_i=[\mathbf Z_{i,1},\ldots,\mathbf Z_{i,T_i}].
\end{equation}
where $T_i$ is the number of policy queries in source episode $i$.
We organize each sequence into contiguous fragments, forming ordered memory units for subsequent retrieval, recomposition, and context refinement,
\begin{equation}
    \mathcal Z_i
    = \mathcal F_i^1\circ\mathcal F_i^2\circ\cdots\circ\mathcal F_i^{K_i},
\end{equation}
where $K_i$ is the number of fragments in episode $i$, $\circ$ denotes temporal
concatenation, and each $\mathcal F_i^k$ contains one or more latent atoms from a
local stage of the interaction. The partition follows the ordered interaction
structure available during memory construction.

Each fragment is stored as
\begin{equation}
    m_i^k=
    (\mathcal F_i^k,\mathbf d_i^k,\bar{\mathbf s}_i^k,
    [p_{i,k}^{\mathrm{start}},p_{i,k}^{\mathrm{end}}],
    \mathbf c_i^k,r_i^k),
\end{equation}
where $\mathbf d_i^k$ and $\bar{\mathbf s}_i^k$ are the observable descriptor and normalized state at the fragment anchor, $p_{i,k}$ records relative progress, $\mathbf c_i^k$ summarizes available transition information, and $r_i^k$ records source provenance and order. Thus, the memory bank contains latent payloads and the minimum metadata required to retrieve and recompose them.

\subsection{Recompose-and-Refine Reasoning Expert}
\label{sec:r2_reasoner}
The Recompose-and-Refine Reasoning Expert maps the current query and memory bank to a fixed-size latent reasoning state. Its three stages correspond to fragment alignment, route tokenization, and context refinement.

\paragraph{Fragment alignment.}
We first restrict retrieval to the task or instruction namespace associated with $l$ and exclude the currently evaluated episode. For each eligible fragment, the retrieval score combines visual semantic similarity with normalized state compatibility:
\begin{equation}
    S(\mathbf q_t,m_i^k)
    = \lambda_v\operatorname{cos}(\mathbf d_t,\mathbf d_i^k)
    -\frac{\lambda_s}{d_s}\left\|
    \frac{\bar{\mathbf s}_t-\bar{\mathbf s}_i^k}{\boldsymbol\sigma_s}
    \right\|_2^2,
\end{equation}
where $d_s$ is the state-vector dimension, $\boldsymbol\sigma_s$ is estimated
from the indexed states, and $\lambda_v,\lambda_s\geq0$. A high-scoring fragment
identifies an anchor inside a successful source trajectory. Rather than retrieving
only that point, we retain the ordered local suffix beginning at the anchor. This
preserves the transition structure that follows a state compatible with the
current scene.

The retained suffixes from several source episodes form the candidate set. We compose them with a deterministic constrained search that preserves within-source order, rejects repeated fragments, and requires consecutive fragments to be compatible in observable state, progress, and any available transition metadata. The selected route is
\begin{equation}
    \mathcal R_t=(m_{i_1}^{k_1},m_{i_2}^{k_2},\ldots,m_{i_J}^{k_J}),
\end{equation}
where $J$ is the number of selected fragments and adjacent elements may originate
from different successful episodes. Only latent reasoning fragments, rather than
historical action segments, participate in route recomposition.

\paragraph{Route tokenization.}
The route $\mathcal R_t$ has a variable number of fragments and latent atoms. For each atom, we add embeddings of its robot state, progress interval, source identity, route position, and transition metadata to the latent payload. Let $\widetilde{\mathcal R}_t$ denote the resulting ordered token sequence. A bank of $L$ learned queries attends to this sequence and produces a fixed-size representation:
\begin{equation}
    \mathbf Z_t^0
    = C_\theta(\widetilde{\mathcal R}_t)
    \in\mathbb R^{L\times d}.
\end{equation}
where $C_\theta$ denotes the learned-query route-tokenization module. This
operation retains source and temporal organization while converting routes of
different lengths into the latent interface expected by the policy.
When no compatible route is available, the policy falls back to the backbone's
native latent-reasoning path.

\paragraph{Context refinement.}
Retrieved reasoning supplies a long-horizon computational prior, whose query-local route tokens are refined in light of current embodied evidence. We formulate this evidence-driven route-token refinement through a fixed-depth residual operator over the route tokens:
\begin{equation}
    \mathbf Z_t'
    = \mathbf Z_t^0
    + \Delta_\theta(\mathbf Z_t^0,\mathbf H_t,\bar{\mathbf s}_t).
\end{equation}
Here, $\Delta_\theta$ is a fixed-depth residual operator that derives query
features from the retrieved route and state, reads the current context as
key-value evidence, and produces a residual correction to the query-local route
tokens.

\subsection{Memory-Adaptive Action Expert}
\label{sec:action_expert}
The refined tokens $\mathbf Z_t'$ are mapped into the VLA backbone's native latent-conditioning interface and combined with the current context. We denote this interface operation by
\begin{equation}
    \widetilde{\mathbf H}_t
    = I_\theta(\mathbf H_t,\mathbf Z_t').
\end{equation}
where $I_\theta$ denotes the native latent-conditioning interface. The action
expert then predicts a conditional velocity field from $\mathbf a^\tau$ at flow
time $\tau$, where $\mathbf a^\tau=\tau\mathbf a^0+(1-\tau)\mathbf a^1$,
\begin{equation}
    \mathbf v_\theta
    = A_\theta(\mathbf a^\tau,\tau,\widetilde{\mathbf H}_t),
\end{equation}
which is integrated to obtain the action chunk $\hat{\mathbf a}_{t:t+H-1}$. The expert consumes the current embodied context and the refined reusable reasoning state.

\subsection{Learning Objectives}
\label{sec:learning_objectives}
We train the memory pathway to reconstruct the frozen native latent reasoning target associated with the current embodied context. Let $\mathbf Z_t^*$ denote the target reasoning tokens and $\mathbf Z_t'$ the memory-conditioned prediction. Each reasoning token is a $d$-dimensional continuous vector in the VLA's latent reasoning space: $\mathbf Z_t^*$ is the frozen reference sequence and $\mathbf Z_t'$ is its memory-conditioned prediction. The latent alignment objective is
\begin{equation}
    \mathcal L_{\mathrm{latent}}
    = \frac{1}{L}\sum_{j=1}^{L}
    \left(1-\operatorname{cos}
    (\mathbf Z_{t,j}',\operatorname{sg}(\mathbf Z_{t,j}^*))\right),
\end{equation}
where $j$ indexes the $L$ reasoning tokens and $\operatorname{sg}(\cdot)$ stops
gradients through the target. The original action-learning objective of the VLA
is retained; for flow-based action experts it is the conditional flow-matching
loss
\begin{equation}
    \mathcal L_{\mathrm{act}}
    = \mathbb E_{\tau,\mathbf a^0,\mathbf a^1}
    \left\|\mathbf v_\theta(\mathbf a^\tau,\tau,\widetilde{\mathbf H}_t)
    -(\mathbf a^0-\mathbf a^1)\right\|_2^2.
\end{equation}
where $\mathbf a^0$ is sampled noise and $\mathbf a^1$ is the target action
chunk.
The complete objective combines latent alignment with the original action-learning objective:
\begin{equation}
    \mathcal L
    = \mathcal L_{\mathrm{act}}
    +\mathcal L_{\mathrm{latent}}.
\end{equation}
This objective directly supervises reusable latent reasoning and executable actions.

\section{Experiments}
\label{sec:experiments}

\subsection{Experimental Setup}
\label{sec:experimental_setup}

\paragraph{Benchmarks and protocol.}
We evaluate \method on RoboMME \citep{dai2026robomme} and LIBERO-Plus
\citep{fei2026liberoplus}. RoboMME evaluates temporal, spatial, object, and
procedural memory through Counting, Permanence, Reference, and Imitation. We
follow its full-16 protocol, evaluating 50 fixed episodes for each of the 16
tasks and reporting the task-macro success rate. Training, memory-source, and
evaluation episodes are disjoint.

For LIBERO-Plus, every policy is trained only on the corresponding standard
LIBERO suite \citep{liu2023libero} and is then evaluated zero-shot under the
seven perturbation domains: camera viewpoint, robot initialization, language,
lighting, background, observation noise, and layout. The main result reports
success separately for the four original LIBERO suites, while the appendix
breaks each method down by perturbation domain. The perturbed instances are
used exclusively for zero-shot evaluation.

\paragraph{Baselines.}
On RoboMME, we consider no-memory policies, past-action conditioning,
episodic memory, and the symbolic, perceptual, and recurrent MME-VLA variants
\citep{fang2025sam2act,sridhar2026memer,dai2026robomme}. Together, these
baselines contrast memory derived from action history, symbolic experience,
visual context, and recurrent state. On LIBERO-Plus, we compare with
suite-specific OpenVLA, UniVLA, OpenVLA-OFT \citep{kim2025openvlaoft},
MemoryVLA \citep{shi2026memoryvla}, MergeVLA \citep{fu2026mergevla}, and
VLA-Adapter \citep{wang2026vlaadapter}.

Our controlled no-memory baseline, \lastzero \citep{liu2026last0}, reasons
only from the current embodied input and neither constructs nor queries reasoning
memory. All controlled models use the same benchmark-specific split and
evaluation manifest.

\subsection{Main Results}
\label{sec:main_results}

Table~\ref{tab:robomme_main} evaluates whether reusable latent reasoning
improves closed-loop performance on memory-dependent manipulation.
Table~\ref{tab:liberoplus_main} tests whether the same mechanism remains useful
under zero-shot perturbations. Detailed RoboMME task results are deferred to
Appendix~\ref{sec:additional_results}.

\begin{table}[H]
    \centering
    \small
    \setlength{\tabcolsep}{5pt}
    \renewcommand{\arraystretch}{1.05}
    \caption{\textbf{Main results on RoboMME.} Success rate (\%) under the
    official Full-16 protocol. Higher is better. Best and second-best are
    bolded and underlined.}
    \label{tab:robomme_main}
    \resizebox{\textwidth}{!}{%
    \begin{tabular}{lcccccc}
        \toprule
        \textbf{Memory} & \textbf{Method} & \textbf{Counting} & \textbf{Permanence}
        & \textbf{Reference} & \textbf{Imitation} & \textbf{Avg.} \\
        \midrule
        None & $\pi_{0.5}$ & 28.8 & 17.0 & 17.2 & 8.8 & 17.9 \\
        Action history & $\pi_{0.5}$ + past actions & 29.1 & 22.8 & 15.9 & 11.2 & 19.7 \\
        \multirow{2}{*}{Symbolic} & SimpleSG + QwenVL & 44.6 & 19.6 & 25.2 & 26.6 & 29.0 \\
        & GroundSG + QwenVL & 38.0 & \underline{39.3} & 31.6 & 21.9 & 32.7 \\
        Episodic & MemER & 48.8 & \textbf{53.2} & \textbf{38.0} & 29.5 & 42.4 \\
        \multirow{7}{*}{Perceptual} & SAM2Act+ & 35.3 & 26.0 & 16.8 & 7.3 & 21.4 \\
        & TokenDrop + Context & 57.9 & 26.9 & 23.7 & 29.4 & 34.5 \\
        & TokenDrop + Modul & 52.3 & 26.8 & 34.7 & 38.3 & 38.0 \\
        & TokenDrop + Expert & 59.4 & 23.6 & 25.3 & 31.1 & 34.9 \\
        & FrameSamp + Context & 50.1 & 23.3 & 21.0 & 28.3 & 30.7 \\
        & FrameSamp + Modul & 65.2 & 25.1 & 36.3 & \textbf{51.4} & 44.5 \\
        & FrameSamp + Expert & 66.8 & 25.2 & 24.1 & 28.9 & 36.3 \\
        \multirow{2}{*}{Recurrent} & TTT + Modul & 34.5 & 22.2 & 20.4 & 10.7 & 22.0 \\
        & RMT + Modul & 34.1 & 15.8 & 21.1 & 9.7 & 20.2 \\
        \midrule
        None & \lastzero & \underline{71.5} & 28.0 & 36.5 & 49.0 & \underline{46.3} \\
        \rowcolor{gray!15}
        \textbf{Reasoning flow} & \textbf{\method (Ours)} & \textbf{75.0} & 29.0 & \underline{37.5} & \underline{50.5} & \textbf{48.0} \\
        \bottomrule
    \end{tabular}}
\end{table}

\begin{table}[H]
    \centering
    \small
    \setlength{\tabcolsep}{6pt}
    \renewcommand{\arraystretch}{1.08}
    \caption{\textbf{Main results on LIBERO-Plus.} Success rate (\%) following
    suite-specific standard-LIBERO training and zero-shot LIBERO-Plus evaluation.
    \textbf{Avg.} pools all 10,030 instances across the four suites. Higher is
    better. Best and second-best are bolded and underlined.}
    \label{tab:liberoplus_main}
    \begin{tabular}{lcccccc}
        \toprule
        \textbf{Method} & \textbf{Venue} & \textbf{Spatial} & \textbf{Object}
        & \textbf{Goal} & \textbf{Long} & \textbf{Avg.} \\
        \midrule
        OpenVLA & CoRL'24 & 19.4 & 14.0 & 15.1 & 14.3 & 15.6 \\
        UniVLA & RSS'25 & 55.5 & 36.7 & 40.7 & 39.9 & 42.9 \\
        OpenVLA-OFT & RSS'25 & \underline{84.0} & 66.5 & \underline{63.0} & 66.4 & 69.6 \\
        MemoryVLA & ICLR'26 & 64.7 & 52.4 & 50.7 & 52.8 & 55.0 \\
        MergeVLA & CVPR'26 & 83.7 & 80.4 & \textbf{65.6} & 59.0 & 72.0 \\
        VLA-Adapter & AAAI'26 & 44.7 & 40.8 & 43.6 & 47.9 & 44.2 \\
        \midrule
        \lastzero & ICML'26 & 72.8 & \underline{83.7} & 62.6 & \underline{73.8} & \underline{73.2} \\
        \rowcolor{gray!15}
        \textbf{\method (Ours)} & -- & \textbf{84.1} & \textbf{84.0} & 62.9 & \textbf{78.7} & \textbf{77.3} \\
        \bottomrule
    \end{tabular}
\end{table}

\paragraph{Main findings.}
Tables~\ref{tab:robomme_main} and~\ref{tab:liberoplus_main} show that
\method achieves the highest aggregate result among the compared methods on
both benchmarks. On RoboMME,
it improves the Full-16 average from 46.3\% to 48.0\% over the controlled
no-memory policy, with gains in Counting (75.0\% vs.\ 71.5\%), Reference
(37.5\% vs.\ 36.5\%), Imitation (50.5\% vs.\ 49.0\%), and Permanence
(29.0\% vs.\ 28.0\%). On LIBERO-Plus, \method raises the pooled success rate
from 73.2\% to 77.3\%. The largest gains occur under spatial and long-horizon
perturbations, improving Spatial from 72.8\% to 84.1\% and Long from 73.8\%
to 78.7\%, while retaining gains on Object and Goal. Together with the
controlled analyses below, these results are consistent with reusable reasoning
flows supplying transferable prior computation and current-context refinement
supporting their use under closed-loop task variation and zero-shot distribution
shifts.

\subsection{Evidence for Latent-Reasoning Reuse}
\label{sec:reuse_evidence}

\paragraph{Compositional reuse.}
On the full RoboMME protocol, we use the same memory bank as in
Table~\ref{tab:robomme_main} and vary how latent reasoning is retrieved and
recomposed. The target episodes, policy interface, and action budget remain
fixed. Table~\ref{tab:reuse} contrasts nearest-state and single-source
retrieval with ordered multi-source reuse.
Nearest-state retrieval reaches 45.6\%, below the 46.3\% no-memory policy,
showing that local latent similarity alone is insufficient. Restricting
retrieval to one source reduces valid-route coverage from 100.0\% to 89.1\%
and success to 44.0\%. In contrast, \method composes compatible fragments across
sources for 70.5\% of policy queries and reaches 48.0\%. Thus, multi-source
recomposition expands the available compatible reasoning evidence and is
associated with the strongest closed-loop performance. The same direction holds
on LIBERO-Plus: Table~\ref{tab:ablation} improves from 76.5\% with a single
source to 77.3\% with the full model.
\begin{center}
	\begin{minipage}[t]{0.48\columnwidth}
		\footnotesize
		\setlength{\tabcolsep}{1.5pt}
		\renewcommand{\arraystretch}{1.05}
		\captionof{table}{\textbf{Compositional reuse on RoboMME.} Memory-conditioned
			methods use the same bank. Higher SR is better. Best and second-best are
			bolded and underlined.}
		\label{tab:reuse}
		\vspace{-2pt}
		\begin{tabular*}{\linewidth}{@{\extracolsep{\fill}}p{0.43\linewidth}ccc@{}}
			\toprule
			\textbf{Method} & \textbf{Valid} & \textbf{Multi} & \textbf{SR} \\
			\midrule
			\lastzero & -- & -- & \underline{46.3} \\
			Nearest state & -- & -- & 45.6 \\
			Single source & 89.1 & 0.0 & 44.0 \\
			w/o refinement & 100.0 & 69.4 & \underline{46.3} \\
			\rowcolor{gray!15}
			\textbf{\method (Ours)} & 100.0 & 70.5 & \textbf{48.0} \\
			\bottomrule
		\end{tabular*}
	\end{minipage}\hfill
	\begin{minipage}[t]{0.48\columnwidth}
		\footnotesize
		\setlength{\tabcolsep}{1.5pt}
		\renewcommand{\arraystretch}{1.05}
		\captionof{table}{\textbf{Memory counterfactuals on RoboMME.}
			$\checkmark$: retained property; blank: corrupted. Higher SR is better.
			Best and second-best are bolded and underlined.}
		\label{tab:counterfactual}
		\vspace{-2pt}
		\begin{tabular*}{\linewidth}{@{\extracolsep{\fill}}p{0.37\linewidth}cccc@{}}
			\toprule
			\textbf{Method} & \textbf{Task} & \textbf{Order} & \textbf{Prog.} & \textbf{SR} \\
			\midrule
			\lastzero & -- & -- & -- & \underline{46.3} \\
			Wrong task & & \checkmark & & 46.1 \\
			Out of order & \checkmark & & & 44.4 \\
			Wrong progress & \checkmark & \checkmark & & 46.1 \\
			\rowcolor{gray!15}
			\textbf{Correct memory} & \checkmark & \checkmark & \checkmark & \textbf{48.0} \\
			\bottomrule
		\end{tabular*}
	\end{minipage}
\end{center}

\paragraph{Memory counterfactuals.}
On the same RoboMME Full-16 manifest and memory bank as Table~\ref{tab:reuse},
the memory-conditioned variants share a checkpoint and embodied input; only
the preserved properties of the retrieved memory change. The memory-free
\lastzero\ row uses its own checkpoint. Table~\ref{tab:counterfactual} tests
whether performance depends on task identity, temporal order, and progress
alignment rather than the presence of arbitrary extra context.
Correct memory attains 48.0\%, exceeding wrong-task and wrong-progress memory
by 1.9 points and out-of-order memory by 3.6 points. The largest drop follows
disruption of temporal order, while the remaining counterfactuals are consistent
with task identity and progress alignment contributing to reliable reasoning
reuse. Under this protocol, the pattern is inconsistent with the explanation
that arbitrary additional latent context is equally useful.

\paragraph{Core ablations.}
Table~\ref{tab:ablation} isolates the three proposed ingredients: representing
experience as an ordered flow, recomposing fragments across episodes, and
refining the resulting route with the current observation. All variants retain
the same action interface and latent-token budget. On RoboMME, the
memory-conditioned variants use the same checkpoint. On LIBERO-Plus, the
single-source variant is trained separately from the same LaST0 base with the
corresponding standard-LIBERO suite and matched training budget.

\begin{table}[!htbp]
    \centering
    \footnotesize
    \setlength{\tabcolsep}{2.3pt}
    \renewcommand{\arraystretch}{1.05}
    \caption{\textbf{Core ablation of FlowMem.} Variants share the same action
    interface and latent-token budget. Higher is better. Best and second-best
    are bolded and underlined.}
    \label{tab:ablation}
    \begin{tabular}{p{0.34\columnwidth}ccccc}
        \toprule
        \textbf{Method} & \textbf{Flow} & \textbf{Multi-src.} & \textbf{Refine}
        & \textbf{RoboMME} & \textbf{LIBERO-Plus} \\
        \midrule
        \lastzero & & & & \underline{46.3} & 73.2 \\
        Single-source FlowMem & \checkmark & & \checkmark & 44.0 & 76.5 \\
        FlowMem w/o refinement & \checkmark & \checkmark & & \underline{46.3} & \underline{76.7} \\
        \rowcolor{gray!15}
        \textbf{\method (Ours)} & \checkmark & \checkmark & \checkmark & \textbf{48.0} & \textbf{77.3} \\
        \bottomrule
    \end{tabular}
\end{table}

\FloatBarrier
The ablations are consistent with benefits from the full design on both
benchmarks. On RoboMME, removing multi-source recomposition reduces success from
48.0\% to 44.0\%, while
removing current-context refinement reduces it to 46.3\%. The same ordering
appears on the complete 10,030-instance LIBERO-Plus evaluation: the
single-source and no-refinement variants obtain 76.5\% and 76.7\%, respectively,
below the full model's 77.3\% but above the 73.2\% no-memory baseline. Thus, on
these benchmark protocols, multi-source recomposition is associated with gains
beyond RoboMME, and current-observation refinement with a further, consistent
improvement when the retrieved route is deployed under closed-loop variation.

\section{Related Work}
\label{sec:related-work}

Recent vision-language-action models have introduced intermediate reasoning between multimodal perception and action generation. ECoT generates embodied reasoning about plans, subgoals, motions, and visually grounded entities, whereas CoT-VLA predicts future images as visual goals before producing actions \citep{zawalski2024ecot,zhao2025cotvla}. ThinkAct compresses action-aligned planning into a visual latent representation, and Fast-ThinkAct further distills lengthy reasoning traces into a small number of continuous latent tokens for efficient control \citep{huang2025thinkact,huang2026fastthinkact}. More recent approaches internalize reasoning directly in continuous representation spaces. LaRA-VLA combines textual reasoning and future-oriented visual prediction in latent reasoning slots, while LaST$_0$ learns a spatio-temporal latent chain-of-thought supervised by future visual dynamics, 3D structure, and robot proprioception \citep{bai2026laravla,liu2026last0}. RD-VLA instead performs weight-tied recurrent refinement to adapt the amount of latent computation at test time \citep{tur2026rdvla}. Complementary geometric analysis models reasoning as a context-cumulative flow in representation space and shows that its velocity, curvature, and ordering capture structure beyond individual embedding positions \citep{zhou2026geometryreasoning}. This perspective motivates characterizing reasoning through its ordered evolution rather than as exchangeable latent vectors. Existing reasoning-enhanced VLAs nevertheless produce or refine such intermediate computation only for the current input: successful latent reasoning trajectories are not retained as experience that can initialize and guide future reasoning.

A separate line of work equips robot policies with memory to address history-dependent control. Notes-to-Self maintains a structured textual scratchpad that records grounding, plans, and completed actions, while MemoryVLA retrieves low-level perceptual details and high-level cognitive semantics from a temporal memory bank. MemoryVLA++ further combines temporal modeling with memory and imagination \citep{haresh2026notestoself,shi2026memoryvla,shi2026memoryvlapp}. Recurrent approaches such as $\mu$VLA, ReMem-VLA, and TFP instead propagate compact memory tokens, recurrent queries, or latent belief states across environment steps \citep{cherepanov2026muvla,li2026rememvla,liang2026tfp}, whereas MemER retrieves bounded visual experience for temporally dependent control \citep{sridhar2026memer}. RoboMME systematizes these requirements into temporal, spatial, object, and procedural memory and evaluates multiple memory representations and integration strategies under a shared $\pi_{0.5}$ backbone \citep{dai2026robomme}. These methods demonstrate that historical context is essential for non-Markovian manipulation. Their memories, however, represent textual task state, perceptual-cognitive evidence, retrieved observations, or recurrent policy state rather than the latent reasoning process that connects embodied observations to action generation.

Several recent approaches lie near the intersection of reasoning and memory, but expose different interfaces. TRM-VLA retrieves explicit reasoning history at key decision points to maintain temporal consistency within an episode \citep{li2026trmvla}. LaMem-VLA retrieves visual and action history, condenses it into latent memory tokens, and injects them as policy context \citep{qu2026lamemvla}. OptimusVLA retrieves trajectory-level action priors and combines them with local execution history to improve action generation \citep{li2026optimusvla}, while WeaveLA transfers a compressed latent state across event-defined subtask boundaries \citep{zhu2026weavela}. Beyond robotic control, MemGen generates latent memory tokens from experience consolidated in auxiliary parameters and uses them to alter subsequent reasoning in language agents \citep{zhang2026memgen}. Thus, prior work has separately explored explicit reasoning history, latent policy memory, trajectory-conditioned action generation, event-triggered latent state transfer, and memory-conditioned language reasoning. We study their intersection in embodied VLAs: successful latent reasoning trajectories are preserved as ordered, source-grounded flows; compatible local flows are retrieved from multiple episodes and recomposed into a task-conditioned reasoning route; and the resulting route is iteratively refined against the current observation before being decoded by the original action expert.

\section{Conclusion}
\label{sec:conclusion}

We presented \method, a VLA framework that turns successful latent computation
into reusable reasoning-flow memory. \method stores ordered latent fragments
with progress and transition metadata, retrieves anchor-aligned local suffixes,
recomposes compatible multi-source routes, and refines the resulting tokens with
the current embodied context before action generation. Across RoboMME and
LIBERO-Plus, \method improves the reported main-task success over the
memory-free \lastzero baseline. Controlled retrieval and memory-counterfactual
experiments are consistent with source compatibility, temporal order, and
progress alignment contributing to performance. These results support reusable
latent reasoning flows as a practical mechanism for connecting successful prior
computation to closed-loop VLA control.

\subsection*{AI Use Statement}
We used generative AI tools to assist with experiment and methodology planning,
implementation and debugging, analysis and interpretation of experimental
results, and drafting, editing, and formatting parts of the manuscript and
figures. Generative AI was not used to create synthetic datasets or to establish
theoretical or mathematical claims. All experiments were executed using the
reported code and evaluation protocols, and the authors manually reviewed
AI-assisted text, code, figures, and analyses against the underlying
implementations, logs, and cited sources. We take responsibility for the final
content of this work, including all claims and artifacts produced with the aid
of generative AI.


\subsection*{Reproducibility Statement}
Sections~3 and~4 specify the model formulation, training objective, benchmark
protocols, controlled baselines, and ablation settings. Table captions state
the metrics and aggregation rules, and Appendix~\ref{sec:additional_results}
provides per-task RoboMME and perturbation-wise LIBERO-Plus results. All
comparisons use the stated protocols and fixed test episodes, with success
rates aggregated from episode-level records.

\bibliography{references}

@article{zawalski2024ecot,
  title={Robotic Control via Embodied Chain-of-Thought Reasoning},
  author={Zawalski, Michal and Chen, William and Pertsch, Karl and Mees, Oier and Finn, Chelsea and Levine, Sergey},
  journal={arXiv preprint arXiv:2407.08693},
  year={2024}
}

@inproceedings{zhao2025cotvla,
  title={{CoT-VLA}: Visual Chain-of-Thought Reasoning for Vision-Language-Action Models},
  author={Zhao, Qingqing and Lu, Yao and Kim, Moo Jin and Fu, Zipeng and Zhang, Zhuoyang and Wu, Yecheng and Li, Zhaoshuo and Ma, Qianli and Han, Song and Finn, Chelsea and others},
  booktitle={IEEE/CVF Conference on Computer Vision and Pattern Recognition},
  pages={1702--1713},
  year={2025}
}

@inproceedings{huang2025thinkact,
  title={{ThinkAct}: Vision-Language-Action Reasoning via Reinforced Visual Latent Planning},
  author={Huang, Chi-Pin and Wu, Yueh-Hua and Chen, Min-Hung and Wang, Yu-Chiang Frank and Yang, Fu-En},
  booktitle={Advances in Neural Information Processing Systems},
  year={2025}
}

@inproceedings{huang2026fastthinkact,
  title={{Fast-ThinkAct}: Efficient Vision-Language-Action Reasoning via Verbalizable Latent Planning},
  author={Huang, Chi-Pin and Man, Yunze and Yu, Zhiding and Chen, Min-Hung and Kautz, Jan and Wang, Yu-Chiang Frank and Yang, Fu-En},
  booktitle={IEEE/CVF Conference on Computer Vision and Pattern Recognition},
  pages={5070--5081},
  year={2026}
}

@inproceedings{bai2026laravla,
  title={Latent Reasoning {VLA}: Latent Thinking and Prediction for Vision-Language-Action Models},
  author={Bai, Shuanghao and Lyu, Jing and Zhou, Wanqi and Li, Zhe and Wang, Dakai and Xing, Lei and Zhao, Xiaoguang and Wang, Pengwei and Wang, Zhongyuan and Chi, Cheng and others},
  booktitle={International Conference on Machine Learning},
  year={2026}
}

@inproceedings{liu2026last0,
  title={{LaST$_0$}: Latent Spatio-Temporal Chain-of-Thought for Robotic Vision-Language-Action Model},
  author={Liu, Zhuoyang and Liu, Jiaming and Chen, Hao and Guo, Ziyu and Hou, Chengkai and Gu, Chenyang and Yu, Jiale and Mi, Xiangju and Zhang, Renrui and Che, Zhengping and others},
  booktitle={International Conference on Machine Learning},
  year={2026}
}

@article{tur2026rdvla,
  title={Recurrent-Depth {VLA}: Implicit Test-Time Compute Scaling of Vision-Language-Action Models via Latent Iterative Reasoning},
  author={Tur, Yalcin and Naghiyev, Jalal and Fang, Haoquan and Tsai, Wei-Chuan and Duan, Jiafei and Fox, Dieter and Krishna, Ranjay},
  journal={arXiv preprint arXiv:2602.07845},
  year={2026}
}

@inproceedings{zhou2026geometryreasoning,
  title={The Geometry of Reasoning: Flowing Logics in Representation Space},
  author={Zhou, Yufa and Wang, Yixiao and Yin, Xunjian and Zhou, Shuyan and Zhang, Anru R.},
  booktitle={International Conference on Learning Representations},
  year={2026}
}

@inproceedings{haresh2026notestoself,
  title={Notes-to-Self: Scratchpad Augmented {VLA}s for Memory Dependent Manipulation Tasks},
  author={Haresh, Sanjay and Dijkman, Daniel and Bhattacharyya, Apratim and Memisevic, Roland},
  booktitle={IEEE International Conference on Robotics and Automation},
  year={2026}
}

@inproceedings{shi2026memoryvla,
  title={{MemoryVLA}: Perceptual-Cognitive Memory in Vision-Language-Action Models for Robotic Manipulation},
  author={Shi, Hao and Xie, Bin and Liu, Yingfei and Sun, Lin and Liu, Fengrong and Wang, Tiancai and Zhou, Erjin and Fan, Haoqiang and Zhang, Xiangyu and Huang, Gao},
  booktitle={International Conference on Learning Representations},
  year={2026}
}

@article{cherepanov2026muvla,
  title={$\mu${VLA}: On Recurrent Memory for Partially Observable Manipulation in {VLA} Models},
  author={Cherepanov, Egor and Kachaev, Nikita and Zelezetsky, Daniil and Bulatov, Aydar and Pshenitsyn, Artem and Kuratov, Yuri and Skrynnik, Alexey and Panov, Aleksandr I. and Kovalev, Alexey K.},
  journal={arXiv preprint arXiv:2606.12497},
  year={2026}
}

@article{li2026rememvla,
  title={{ReMem-VLA}: Empowering Vision-Language-Action Model with Memory via Dual-Level Recurrent Queries},
  author={Li, Hang and Shen, Fengyi and Chen, Dong and Yang, Liudi and Wang, Xudong and Shi, Jinkui and Bing, Zhenshan and Liu, Ziyuan and Knoll, Alois},
  journal={arXiv preprint arXiv:2603.12942},
  year={2026}
}

@article{liang2026tfp,
  title={{TFP}: Temporally Conditioned Memory-Fusion Policies for Visuomotor Learning},
  author={Liang, Yushen and Peng, Yue and Jin, Baosheng and Zhang, Tianluo and Zhang, Xinyu and Zhou, Shuyi and Chen, Zhuoran and Liu, Xinqi and Wan, Shenji},
  journal={arXiv preprint arXiv:2607.08283},
  year={2026}
}

@inproceedings{sridhar2026memer,
  title={{MemER}: Scaling Up Memory for Robotic Control via Experience Retrieval},
  author={Sridhar, Ajay and Pan, Jennifer and Sharma, Satvik and Finn, Chelsea},
  booktitle={International Conference on Learning Representations},
  year={2026}
}

@inproceedings{dai2026robomme,
  title={{RoboMME}: Benchmarking and Understanding Memory for Robotic Generalist Policies},
  author={Dai, Yinpei and Fu, Hongze and Lee, Jayjun and Liu, Yuejiang and Zhang, Haoran and Yang, Jianing and Finn, Chelsea and Fazeli, Nima and Chai, Joyce},
  booktitle={International Conference on Machine Learning},
  year={2026}
}

@article{fang2025sam2act,
  title={{SAM2Act}: Integrating Visual Foundation Model with A Memory Architecture for Robotic Manipulation},
  author={Fang, Haoquan and Grotz, Markus and Pumacay, Wilbert and Wang, Yi Ru and Fox, Dieter and Krishna, Ranjay and Duan, Jiafei},
  journal={arXiv preprint arXiv:2501.18564},
  year={2025}
}

@inproceedings{fu2026mergevla,
  title={{MergeVLA}: Cross-Skill Model Merging Toward a Generalist Vision-Language-Action Agent},
  author={Fu, Yuxia and Zhang, Zhizhen and Zhang, Yuqi and Wang, Zijian and Huang, Zi and Luo, Yadan},
  booktitle={IEEE/CVF Conference on Computer Vision and Pattern Recognition},
  year={2026}
}

@inproceedings{wang2026vlaadapter,
  title={{VLA}-Adapter: An Effective Paradigm for Tiny-Scale Vision-Language-Action Model},
  author={Wang, Yihao and Ding, Pengxiang and Li, Lingxiao and Cui, Can and Ge, Zirui and Tong, Xinyang and Song, Wenxuan and Zhao, Han and Zhao, Wei and Hou, Pengxu and Huang, Siteng and Tang, Yifan and Wang, Wenhui and Zhang, Ru and Liu, Jianyi and Wang, Donglin},
  booktitle={Proceedings of the AAAI Conference on Artificial Intelligence},
  volume={40},
  number={22},
  pages={18638--18646},
  year={2026}
}

@inproceedings{liu2023libero,
  title={{LIBERO}: Benchmarking Knowledge Transfer for Lifelong Robot Learning},
  author={Liu, Bo and Zhu, Yifeng and Gao, Chongkai and Feng, Yihao and Liu, Qiang and Zhu, Yuke and Stone, Peter},
  booktitle={Advances in Neural Information Processing Systems},
  volume={36},
  year={2023}
}

@inproceedings{fei2026liberoplus,
  title={{LIBERO-Plus}: A Progressive Robustness Benchmark for Visual-Language-Action Models},
  author={Fei, Senyu and Wang, Siyin and Shi, Junhao and Dai, Zihao and Cai, Jikun and Qian, Pengfang and Ji, Li and He, Xinzhe and Zhang, Shiduo and Fei, Zhaoye and Fu, Jinlan and Gong, Jingjing and Qiu, Xipeng},
  booktitle={IEEE/CVF Conference on Computer Vision and Pattern Recognition},
  pages={38574--38583},
  year={2026}
}

@inproceedings{kim2025openvlaoft,
  title={Fine-Tuning Vision-Language-Action Models: Optimizing Speed and Success},
  author={Kim, Moo Jin and Finn, Chelsea and Liang, Percy},
  booktitle={Robotics: Science and Systems},
  year={2025}
}

@article{shi2026memoryvlapp,
  title={{MemoryVLA++}: Temporal Modeling via Memory and Imagination in Vision-Language-Action Models},
  author={Shi, Hao and Li, Weiye and Xie, Bin and Wang, Yulin and Zhou, Renping and Wang, Tiancai and Zhang, Xiangyu and Luo, Ping and Huang, Gao},
  journal={arXiv preprint arXiv:2606.09827},
  year={2026}
}

@inproceedings{li2026trmvla,
  title={{TRM-VLA}: Temporal-Aware Chain-of-Thought Reasoning and Memorization for Vision-Language-Action Models},
  author={Li, Xiang and Li, Ya-Li and Wang, Yuan and Wang, Shengjin},
  booktitle={IEEE/CVF Conference on Computer Vision and Pattern Recognition},
  year={2026}
}

@article{qu2026lamemvla,
  title={Dual Latent Memory in Vision-Language-Action Models for Robotic Manipulation},
  author={Qu, Hongyu and Gao, Jianzhe and Hu, Xiaobin and Yang, Shaohuan and Yu, Xinlei and Yan, Rui and Wang, Wenguan and Shu, Xiangbo and Yan, Shuicheng},
  journal={arXiv preprint arXiv:2607.07608},
  year={2026}
}

@inproceedings{li2026optimusvla,
  title={Global Prior Meets Local Consistency: Dual-Memory Augmented Vision-Language-Action Model for Efficient Robotic Manipulation},
  author={Li, Zaijing and Hu, Bing and Shao, Rui and Chen, Gongwei and Jiang, Dongmei and Xie, Pengwei and Hao, Jianye and Nie, Liqiang},
  booktitle={IEEE/CVF Conference on Computer Vision and Pattern Recognition},
  pages={35135--35145},
  year={2026}
}

@article{zhu2026weavela,
  title={{WeaveLA}: Event Driven Cross-Subtask Latent Memory Weaving for Repetitive Robot Manipulation},
  author={Zhu, Shoujing and Liu, Zhenyang and Wang, Fungmiu and Wang, Jiafeng and Yue, Bo and Liu, Guiliang and Wu, Simo and Xue, Xiangyang and Zeng, Taiping},
  journal={arXiv preprint arXiv:2606.17463},
  year={2026}
}

@inproceedings{zhang2026memgen,
  title={{MemGen}: Weaving Generative Latent Memory for Self-Evolving Agents},
  author={Zhang, Guibin and Fu, Muxin and Yan, Shuicheng},
  booktitle={International Conference on Learning Representations},
  year={2026}
}
\bibliographystyle{flowmem_arxiv}

\appendix
\section{Additional Experimental Results}
\label{sec:additional_results}

\subsection{RoboMME Per-Task Results}

\begin{table}[!htbp]
    \centering
    \small
    \setlength{\tabcolsep}{4pt}
    \renewcommand{\arraystretch}{1.05}
    \caption{\textbf{Per-task RoboMME results.} Success rate (\%) under the
    official 16-task, 50-episode-per-task protocol. Higher is better. Best
    results are bolded.}
    \label{tab:robomme_per_task}
    \begin{tabular*}{\textwidth}{@{\extracolsep{\fill}}l*{8}{c}@{}}
        \toprule
        \textbf{Method} & \rotatebox{60}{\textbf{BinFill}} & \rotatebox{60}{\textbf{PickX}} & \rotatebox{60}{\textbf{SwingX}}
        & \rotatebox{60}{\textbf{StopCube}} & \rotatebox{60}{\textbf{V-Unmask}} & \rotatebox{60}{\textbf{V-Unmask-S}}
        & \rotatebox{60}{\textbf{B-Unmask}} & \rotatebox{60}{\textbf{B-Unmask-S}} \\
        \midrule
        FrameSamp+Modul & 39.6 & 87.3 & 92.0 & 42.0 & 32.7 & 24.4 & 25.1 & 18.2 \\
        \lastzero & 46.0 & 96.0 & 90.0 & 54.0 & 34.0 & 24.0 & \textbf{26.0} & \textbf{28.0} \\
        \rowcolor{gray!15}
        \textbf{\method (Ours)} & \textbf{50.0} & \textbf{98.0} & \textbf{96.0} & \textbf{56.0} & \textbf{40.0} & \textbf{26.0} & 24.0 & 26.0 \\
        \bottomrule
    \end{tabular*}

    \vspace{4pt}
    \begin{tabular*}{\textwidth}{@{\extracolsep{\fill}}l*{9}{c}@{}}
        \toprule
        \textbf{Method} & \rotatebox{60}{\textbf{PickHigh}} & \rotatebox{60}{\textbf{V-Repick}}
        & \rotatebox{60}{\textbf{V-PlaceB}} & \rotatebox{60}{\textbf{V-PlaceO}}
        & \rotatebox{60}{\textbf{MoveCube}} & \rotatebox{60}{\textbf{InsertPeg}}
        & \rotatebox{60}{\textbf{PatternLock}} & \rotatebox{60}{\textbf{RouteStick}} & \textbf{Avg.} \\
        \midrule
        FrameSamp+Modul & 22.9 & \textbf{30.4} & \textbf{60.0} & 32.0 & 77.8 & \textbf{7.6} & 53.6 & \textbf{66.7} & 44.5 \\
        \lastzero & 22.0 & 22.0 & \textbf{60.0} & \textbf{42.0} & 80.0 & 2.0 & 52.0 & 62.0 & 46.3 \\
        \rowcolor{gray!15}
        \textbf{\method (Ours)} & \textbf{24.0} & 26.0 & \textbf{60.0} & 40.0 & \textbf{82.0} & 2.0 & \textbf{56.0} & 62.0 & \textbf{48.0} \\
        \bottomrule
    \end{tabular*}
\end{table}

\FloatBarrier
\paragraph{Per-task analysis.}
The 1.7-point Full-16 gain is distributed across 10 of the 16 tasks (34
positive points versus 6 negative points), rather than being produced by one
outlier. The largest gains occur on Visual Unmask (40.0\% vs.\ 34.0\%), Visual
Repick (26.0\% vs.\ 22.0\%), and Pattern Lock (56.0\% vs.\ 52.0\%); the two
B-Unmask variants and Visual PlaceO are lower, and three tasks are unchanged.
Together with Table~\ref{tab:reuse}, this distribution supports the central
claim that ordered latent reasoning can be reused across distinct closed-loop
tasks, while showing that useful reuse still requires compatibility with the
current embodied context.

\subsection{Detailed LIBERO-Plus Results}

\begin{table}[!htbp]
    \centering
    \footnotesize
    \setlength{\tabcolsep}{3.5pt}
    \renewcommand{\arraystretch}{1.03}
    \caption{\textbf{Perturbation-wise LIBERO-Plus results.} Success rate (\%) following
    suite-specific standard-LIBERO training and zero-shot LIBERO-Plus evaluation.
    Each perturbation column pools its instances across the four suites;
    \textbf{Total} pools all 10,030 instances. Higher is better. Best and
    second-best are bolded and underlined.}
    \label{tab:liberoplus_detailed}
    \begin{tabular}{lccccccccc}
        \toprule
        \textbf{Method} & \textbf{Venue} & \textbf{Camera} & \textbf{Robot}
        & \textbf{Language} & \textbf{Light} & \textbf{Background} & \textbf{Noise}
        & \textbf{Layout} & \textbf{Total} \\
        \midrule
        OpenVLA & CoRL'24 & 0.8 & 3.5 & 23.0 & 8.1 & 34.8 & 15.2 & 28.5 & 15.6 \\
        UniVLA & RSS'25 & 1.8 & \underline{46.2} & 69.6 & 69.0 & 81.0 & 21.2 & 31.9 & 42.9 \\
        OpenVLA-OFT & RSS'25 & 56.4 & 31.9 & 79.5 & 88.7 & \textbf{93.3} & 75.8 & \underline{74.2} & 69.6 \\
        MemoryVLA & ICLR'26 & 10.3 & \textbf{50.2} & 79.5 & 65.3 & 89.4 & 31.1 & \textbf{75.1} & 55.0 \\
        MergeVLA & CVPR'26 & 61.7 & 44.8 & 75.7 & \textbf{92.0} & \underline{93.0} & 73.7 & \textbf{75.1} & 72.0 \\
        VLA-Adapter & AAAI'26 & 6.1 & 29.1 & 66.2 & 56.5 & 70.5 & 25.7 & 69.2 & 44.2 \\
        \midrule
        \lastzero & ICML'26 & \underline{80.2} & 36.3 & \underline{81.8} & 86.5 & 88.8 & \underline{77.5} & 68.8 & \underline{73.2} \\
        \rowcolor{gray!15}
        \textbf{\method} & -- & \textbf{88.4} & 36.6 & \textbf{82.2} & \underline{90.2} & 90.5 & \textbf{88.3} & 71.2 & \textbf{77.3} \\
        \bottomrule
    \end{tabular}
\end{table}

\FloatBarrier
\paragraph{Perturbation-wise analysis.}
The largest gains occur under Camera (80.2\% to 88.4\%) and Noise (77.5\% to
88.3\%), with a further 3.7-point gain under Light. This pattern supports the
central claim: recomposition is useful when stored trajectories provide a
compatible route, while refinement aligns that route with the current
observation rather than treating memory as fixed retrieved context. Gains on
Robot and Language are only 0.3 and 0.4 points, respectively, and \method does
not lead every perturbation column, delimiting where the mechanism offers only
marginal benefit.

\subsection{Inference Cost of Latent-Reasoning Reuse}
\label{sec:efficiency}

Table~\ref{tab:efficiency} reports the mean online control time per episode
under a fixed profiling protocol. We profile 256 episodes per method on an
NVIDIA H800 GPU using the same evaluation configuration. For each episode, we
sum the time from an observation entering the policy to its action chunk being
returned across all control decisions, then average over the profiling episodes.

\begin{table}[!htbp]
    \centering
    \small
    \setlength{\tabcolsep}{4pt}
    \renewcommand{\arraystretch}{1.05}
    \caption{\textbf{Inference efficiency on RoboMME.} \textbf{Time / episode}
    is the mean total online control time from observation to action-chunk return
    over the profiling episodes. Lower time and higher SR are better.}
    \label{tab:efficiency}
    \begin{tabular}{lcc}
        \toprule
        \textbf{Method} & \textbf{Time / episode (s)} & \textbf{SR (\%)} \\
        \midrule
        \lastzero & 19.6 & 46.3 \\
        \method & 15.5 & \textbf{48.0} \\
        \bottomrule
    \end{tabular}
\end{table}

\FloatBarrier
On RoboMME, \method reduces the mean online control time from 19.6 to 15.5
seconds per episode while improving success from 46.3\% to 48.0\%. Under this
profiling protocol, reasoning-flow reuse therefore improves task performance
without increasing online control time, supporting its practical use in
closed-loop control.

\end{document}